\PassOptionsToPackage{sort, compress}{natbib}

\documentclass[11pt, a4paper]{include/gdm_format}
\usepackage[authoryear, sort&compress, round]{natbib}

\usepackage{amsmath,amsfonts,bm}

\def\eqref#1{equation~\ref{#1}}

\def\1{\bm{1}}

\DeclareMathAlphabet{\mathsfit}{\encodingdefault}{\sfdefault}{m}{sl}
\SetMathAlphabet{\mathsfit}{bold}{\encodingdefault}{\sfdefault}{bx}{n}

\usepackage{wrapfig}
\usepackage{hyperref}
\usepackage{url}
\usepackage{booktabs} 
\usepackage{colortbl} 
\usepackage{xcolor} 
\usepackage{array} 
\usepackage{multirow} 
\usepackage{amssymb}
\usepackage{graphicx} 
\usepackage{float}    
\usepackage[most]{tcolorbox}
\usepackage{listings}
\usepackage{fontawesome5}
\usepackage{pgffor}

\usepackage{amsmath}
\usepackage{amssymb} 
\usepackage{xspace}
\usepackage{multirow}
\usepackage{booktabs}
\usepackage{color}
\usepackage{colortbl}
\usepackage{cleveref}
\usepackage{graphicx}
\usepackage{algorithm}
\usepackage{algorithmicx}
\usepackage{algpseudocode}
\usepackage{subcaption}
\usepackage{wrapfig}
\usepackage{sidecap}
\usepackage{soul}
\usepackage{enumitem}
\sidecaptionvpos{figure}{t}
\usepackage{multicol}
\usepackage{pifont}
\usepackage[normalem]{ulem}
\usepackage{titletoc}

\usepackage{CJKutf8}

\definecolor{brightmaroon}{rgb}{0.76, 0.13, 0.28}
\definecolor{dartmouthgreen}{rgb}{0.05, 0.5, 0.06}
\definecolor{darkpastelgreen}{rgb}{0.01, 0.75, 0.24}
\definecolor{darkpastelred}{rgb}{0.76, 0.23, 0.13}

\lstdefinestyle{pythonstyle}{
    language=Python,
    basicstyle=\ttfamily\small,
    keywordstyle=\color{blue},
    commentstyle=\color{green!50!black},
    stringstyle=\color{red},
    showstringspaces=false,
    numbers=left,
    numberstyle=\tiny\color{gray},
    frame=single,
    breaklines=true,
    tabsize=4,
}

\tcbset {
  base/.style={
    arc=0mm, 
    bottomtitle=-0.25mm,
    boxrule=0mm,
    colbacktitle=black!10!white, 
    coltitle=black, 
    fonttitle=\bfseries, 
    left=2.5mm,
    leftrule=1mm,
    right=3.5mm,
    title={#1},
    toptitle=0.25mm,
    breakable,
  }
}

\definecolor{brandblue}{rgb}{0.34, 0.7, 1}
\newtcolorbox{mybox}[1]{
  colframe=brandblue, 
  base={#1}
}

\definecolor{pink}{rgb}{1, 0.75, 0.8}
\newtcolorbox{safetybox}[1]{
  colframe=pink, 
  base={#1}
}

\title{AgentRxiv: Towards Collaborative Autonomous Research}

\correspondingauthor{Samuel Schmidgall (sschmi46@jhu.edu)}

\author[1]{Samuel Schmidgall}
\author[2]{Michael Moor}

\affil[1]{Department of Electrical \& Computer Engineering, Johns Hopkins University}
\affil[2]{Department of Biosystems Science \& Engineering, ETH Zurich}

\begin{document}

\begin{abstract}

Progress in scientific discovery is rarely the result of a single "Eureka" moment, but is rather the product of hundreds of scientists incrementally working together toward a common goal. While existing agent workflows are capable of producing research autonomously, they do so in isolation, without the ability to continuously improve upon prior research results. To address these challenges, we introduce \texttt{AgentRxiv}—a framework that lets LLM agent laboratories upload and retrieve reports from a shared preprint server in order to collaborate, share insights, and iteratively build on each other’s research. We task agent laboratories to develop new reasoning and prompting techniques and find that agents with access to their prior research achieve higher performance improvements compared to agents operating in isolation (11.4\% relative improvement over baseline on MATH-500). We find that the best performing strategy generalizes to benchmarks in other domains (improving on average by 3.3\%). Multiple agent laboratories sharing research through \texttt{AgentRxiv} are able to work together towards a common goal, progressing more rapidly than isolated laboratories, achieving higher overall accuracy (13.7\% relative improvement over baseline on MATH-500). These findings suggest that autonomous agents may play a role in designing future AI systems alongside humans. We hope that \texttt{AgentRxiv} allows agents to collaborate toward research goals and enables researchers to accelerate discovery.

\end{abstract}

\maketitle

\begin{center}

\href{https://AgentRxiv.github.io/}{\faGithub  \xspace \texttt{AgentRxiv.github.io}}

\begin{figure}[ht!]
    \centering
    \includegraphics[width=0.94\linewidth]{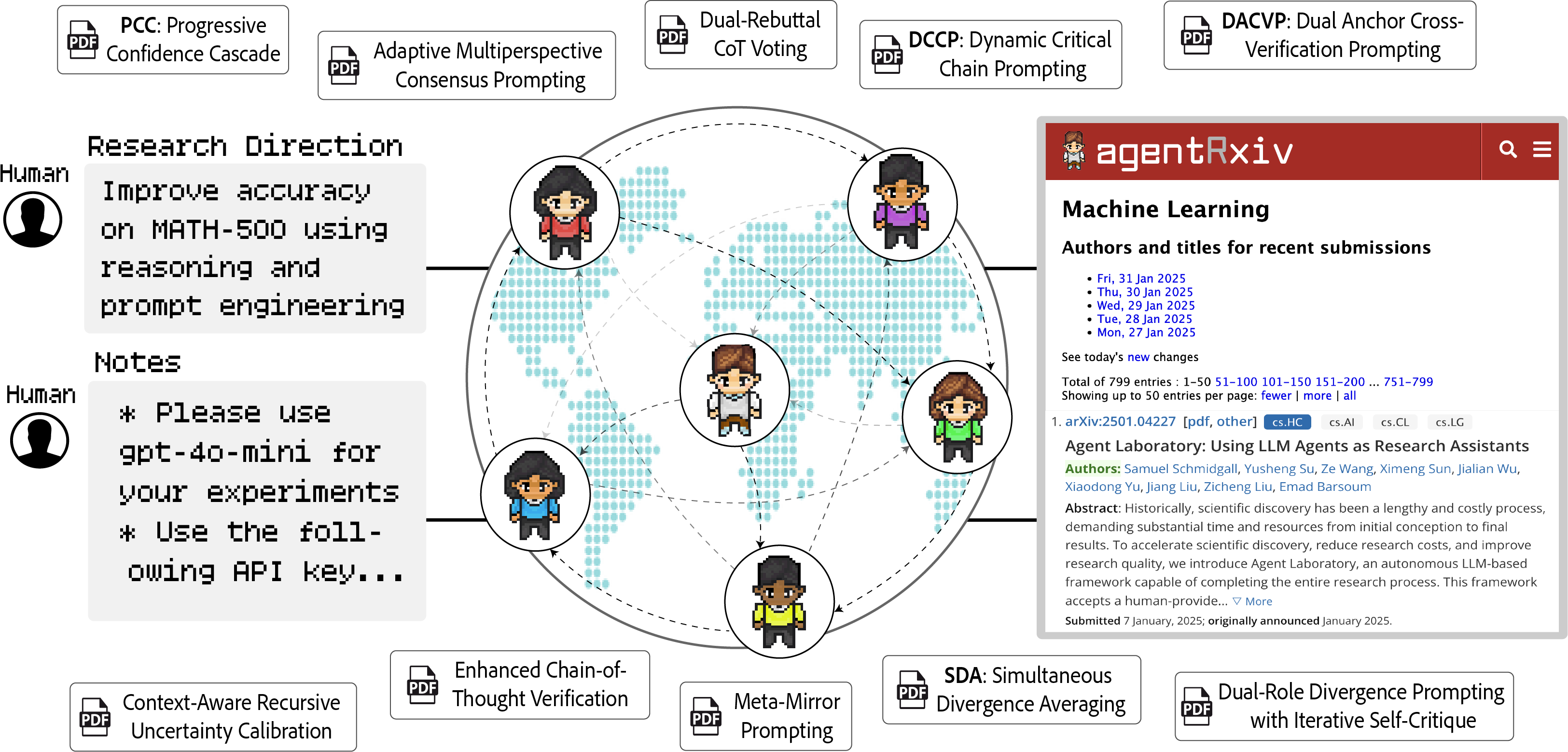}
    \caption{\textbf{Collaborative Autonomous Research via AgentRxiv.} Autonomous agent laboratories distributed collaboratively pursue a shared research goal using \texttt{AgentRxiv}. Human researchers provide initial guidance through a research direction and detailed instructions. Agents autonomously perform research and upload research papers to the centralized \texttt{AgentRxiv} preprint server, enabling laboratories to access each other's discoveries, accelerating scientific progress.}
    \label{fig:SoA}
\end{figure}

\end{center}

\section{Introduction}

Scientific discovery has historically been an iterative process, characterized by the systematic formulation of hypotheses, the execution of controlled experiments, and the evaluation of results (\cite{chalmers2013thing}). Over time, these methods lead to the steady accumulation of knowledge, forming the basis upon which further inquiries are built (\cite{shapere1964structure}). In this manner, scientific progress does not typically arise from isolated breakthroughs but rather from incremental improvements that collectively advance our understanding of complex phenomena.

In an effort to accelerate the process of scientific discovery, recent work has explored the ability of LLM agents to perform autonomous research (\cite{schmidgall2025agent, swanson2024virtual, lu2024aiscientist}).  The AI Scientist framework (\cite{lu2024aiscientist}) is a large language model (LLM)-based system that generates research ideas in machine learning, writes research code, run experiments, and produces a scientific paper with using an automated peer-review to evaluate the work. Virtual Lab (\cite{swanson2024virtual}) uses a multi-agent system of LLM-based experts from different backgrounds (e.g. chemist or biologist) working together with human scientists to produce novel nanobody binders for SARS-CoV-2, where discovered nanobodies demonstrate promising efficacy in wet-lab validations. Finally, \texttt{Agent Laboratory} (\cite{schmidgall2025agent}) is a multi-agent autonomous research system that is able to incorporate human feedback, with greatly reduced cost compared to \cite{lu2024aiscientist}. While these works demonstrate progress toward accelerated scientific discovery, they often operate in \textit{isolation} and do not support the continuous, cumulative development of research across time that reflects the nature of science. Therefore, we aim to provide a unified platform that enables agents to build upon the research of other agents.

In this study, we introduce \texttt{AgentRxiv}, an autonomous research collaboration framework that supports LLM agents in generating, sharing, and building upon scientific research. By implementing a centralized, open-source preprint server for autonomous agents, \texttt{AgentRxiv} enables systematic sharing of research findings, allowing agents to cumulatively build on previous work. \texttt{AgentRxiv} also supports parallel research across multiple agentic systems, enabling scalability with available computational resources. When incorporating \texttt{AgentRxiv}, each generation of papers shows measurable improvements. For example, accuracy on the MATH-500 benchmark increased from 70.2\% to 78.2\% with the best discovered reasoning technique, using gpt-4o mini as the base model. Moreover, reasoning strategies developed on MATH-500 are shown to generalize to other tasks and language models, improving performance on benchmarks such as GPQA, MMLU-Pro, and MedQA across models ranging from DeepSeek-v3 to Gemini-2.0 pro. Although the parallelized mode accelerates improvements in wall-clock time, it introduces trade-offs between speed and computational efficiency. A detailed list of our contributions are provided below:

\begin{itemize}
    \item We introduce \texttt{AgentRxiv}—a novel, open-source framework designed to archive and disseminate research outputs from autonomous agents. This platform enables agents to build upon the discoveries of other agents, driving iterative improvements over time.
    \item We show that each generation of papers produces measurable improvements when agents are given access to \texttt{AgentRxiv}. For instance, accuracy on the MATH-500 benchmark steadily increased from a 70.2\% baseline to 78.2\% (11.4\% relative improvement) using newly discovered techniques like Simultaneous Divergence Averaging (SDA).
    \item We demonstrate that reasoning strategies discovered through \texttt{AgentRxiv} on MATH-500 generalize to other benchmarks and language models. Using the highest performing discovered reasoning method by \texttt{AgentRxiv}, our experiments show performance increase across a range of tasks—from GPQA and MMLU-Pro to MedQA—while also revealing consistent gains across five different language models, from DeepSeek-v3 to Gemini-2.0 pro (improving on average by 3.3\%).
    \item We introduce a parallelized mode for \texttt{AgentRxiv}, allowing multiple agentic systems to run simultaneously and share findings. We show that this setup accelerates improvements on MATH-500 by +6.0\% with 3 parallel labs. We also find that there are a trade-offs between speed and computational efficiency, with discoveries occurring faster but at a higher computational cost.

\end{itemize}

\section{Background and Related Work}

\paragraph{Large language models.} Large language models (LLMs) typically employ transformer architectures (\cite{vaswani2017attention}), which rely on self-attention to capture long-range dependencies in text without using recurrent (\cite{rumelhart1985learning, jordan1997serial}) or convolutional layers (\cite{fukushima1980neocognitron, lecun1998gradient}). These models can contain hundreds of millions or even billions of parameters, often trained on large-scale corpora drawn from diverse sources such as web pages, books, and online forums (\cite{radford2018improving, brown2020language, radford2019language, devlin2019bert}). During training, LLMs are often trained to model statistical regularities in the data by predicting the next token in a sequence, which enables them to generate coherent text, answer questions, and support a wide variety of language understanding tasks. Recent reasoning models—such as OpenAI’s o1 (\cite{jaech2024openai}), o3-mini (\cite{o3mini2025}), and DeepSeek’s R1 (\cite{guo2025deepseek})—extend the paradigm of next-token prediction by incorporating chain-of-thought prompting (\cite{wei2022chain}) and reinforcement learning techniques (\cite{shao2024deepseekmath}) to support multi-step problem solving.

\paragraph{LLM Agents.} LLM Agents expand the capabilities of LLMs by integrating structured workflows in order to autonomously perform task execution (\cite{zhang2024aflow, wu2023autogen, yao2023react, yang2024swe, lu2023chameleon, zhuge2024gptswarm}). Rather than only generating text outputs, these agents interact with the external environment, using techniques such as reasoning (\cite{kojima2022large, yao2023tree, hao2023reasoning, wei2022chain}), iterative refinement (\cite{madaan2023self}), self-improvement (\cite{shinn2023reflexion, huang2022large}), and tool usage (\cite{paranjape2023art, qin2023toolllm, hao2024toolkengpt, m2024augmenting, schick2023toolformer, qu2024toolsurvey}) to accomplish this. LLM agents have made incredible progress across a wide range of tasks from medical task solving (\cite{schmidgall2024agentclinic, li2024agent, tu2024towards, shi2024ehragent}), machine learning engineering (\cite{chan2024mle, guo2024ds, huang2024mlagentbench, nathani2025mlgym}), software engineering  (\cite{jimenez2023swe, yang2024swe, wang2024opendevin}) and web tasks (\cite{gur2023real, putta2024agent, deng2024mind2web, shi2017world, he2024webvoyager}).

\paragraph{Automated machine learning.}

Automated machine learning (AutoML) focuses on developing systems that autonomously select models, optimize hyperparameters, and perform feature engineering to improve model performance (\cite{he2021automl, tornede2023automl}). Recently, many works using LLMs in AutoML have focused on problems from the online platform \textit{Kaggle}--which hosts machine learning competitions--as a benchmark for evaluating agent performance. Notable efforts include MLE-Bench (\cite{chan2024mle}), ML-Bench \cite{nathani2025mlgym}, DS-bench (\cite{jing2024dsbench}), and MLAgentBench (\cite{huang2024mlagentbench}) which propose using 75, 8, 74, and 6 Kaggle (or Kaggle-style) challenges respectively as benchmarks to measure the abilities of ML agents in tasks such as data preparation, model development, and submission. Several ML "solvers" which can solve ML challenges have been introduced, such as AIDE (\cite{AIDE}), CodeActAgent (referred to as “OpenHands") (\cite{wang2024opendevin}), and ResearchAgent (referred to as “MLAB") from MLAgentBench (\cite{huang2024mlagentbench}) which automate feature implementation, bug fixing, and code refactoring with a high success rate. \texttt{mle-solver} (\cite{schmidgall2025agent}) is a module that iteratively generates, refines, and evaluates ML code using a cycle of command execution, error repair, and LLM reward-based scoring, and demonstrates SOTA performance on a subset of MLE-Bench. AutoKaggle (\cite{li2024autokaggle}) is a user-centric multi-agent system focused on assisting data scientists toward completing data pipelines whereas Agent K (\cite{grosnit2024large}) focuses on autonomy and demonstrates the ability to solve Kaggle challenges at the human-level with only a URL provided as input.

\paragraph{AI in Scientific Discovery.} AI has been used as a tool to support scientific discovery across various disciplines in recent years. Recently, AI has been used for discovery in mathematics (\cite{romera2024mathematical, davies2021advancing, cornelio2023combining, shojaee2024llm}), material science (\cite{szymanski2023autonomous, pyzer2022accelerating, merchant2023scaling, jia2024llmatdesign, ding2024matexpert}), chemistry (\cite{jumper2021highly, hayes2024simulating, yang2024moose}), algorithm discovery and code optimization (\cite{lange2025ai, fawzi2022discovering}), and biology (\cite{ding2024automating, gottweis2025aicoscientist, abramson2024accurate}). These approaches differ from autonomous research by positioning AI as a tool itself rather than as an autonomous agent performing research.

\paragraph{LLMs for research related tasks.}
LLMs demonstrate strong capabilities in a wide range of research-related tasks, such as code generation (\cite{chen2021evaluating, nijkamp2022codegen}), end-to-end software development (\cite{qian2024chatdev, qian2023experiential, phan2024hyperagent, hai2024repoexec}), code generation for discovery (\cite{majumder2024discoverybench, ifargan2024autonomous, hu2024infiagent, guo2024ds, gu2024blade, ghafarollahi2024protagents, chen2024scienceagentbench}), research question-answering (\cite{chen2024scholarchemqa, lala2023paperqa, song2024cs, lin2024biokgbench}), research ideation (\cite{ghafarollahi2024sciagents, li2024chain, si2024can, baek2024researchagent}), automated paper reviewing (\cite{d2024marg, liang2024can, lu2024aiscientist, weng2024cycleresearcher, baek2024researchagent}), literature search (\cite{ajith2024litsearch, kang2024researcharena, press2024citeme, li2024scilitllm, baek2024researchagent, gottweis2025aicoscientist}), experiment design (\cite{baek2024researchagent}), and predicting the outcome of experiments (\cite{luo2024large, ashokkumar2024predicting, lehr2024chatgpt, manning2024automated, zhang2024massw}). Notable among these, CycleResearcher (\cite{weng2024cycleresearcher}) introduces two LLMs, CycleReviewer which is trained to predict paper scores from OpenReview, and CycleResearcher which is trained to write high quality research papers. The AI Co-Scientist (\cite{gottweis2025aicoscientist}) is a multi-agent system that helps scientists generate novel hypotheses and research proposals. The hypotheses of this work were validated in real biomedical applications, demonstrating great promise for automated discovery.

\paragraph{LLMs for autonomous research.}

In addition to performing research related tasks, LLMs have also been used to perform end-to-end research. \texttt{Agent Laboratory} (\cite{schmidgall2025agent}) integrates human feedback into a multi-stage LLM agent pipeline--covering literature review, experimentation, and report writing--to produce complete research outputs at reduced cost and improved efficiency compared with other frameworks. \cite{swanson2024virtual} introduces a multi-agent system of LLM scientists  working together with human researchers to produce novel nanobody binders that address recent variants of SARS-CoV-2. The AI Scientist (\cite{lu2024ai}) performs end-to-end discovery in machine learning, including ML coding, paper writing, and automated peer review.  LUMI-lab introduces an active-learning experimental workflow for the design and synthesis of lipid nanoparticles (\cite{Cui2025.02.14.638383}). ChemCrow (\cite{m2024augmenting}) and Coscientist (\cite{boiko2023autonomous}) demonstrate the ability for autonomous research in chemistry.
Curie improves reliability, methodical control, and interpretability during scientific experimentation (\cite{kon2025curie}) and \cite{huang2025automatedhypothesisvalidationagentic} implements sequential falsification and rigorous statistical error control to automatically validate free-form hypotheses at scale.
DISCOVERYWORLD (\cite{jansen2024discoveryworld}) is a virtual, text-based environment where agents perform a range of tasks--from hypothesis formulation and experimental design to execution and analysis--across multiple domains, rather than operating within predefined task constraints. However, the work of \cite{si2024can} demonstrates challenges in the feasibility of LLM-produced research plans, indicating a complementary rather than replacement role for LLMs in autonomous research. 

\subsection{Agent Laboratory}

\begin{figure}
    \centering
    \includegraphics[width=0.98\linewidth]{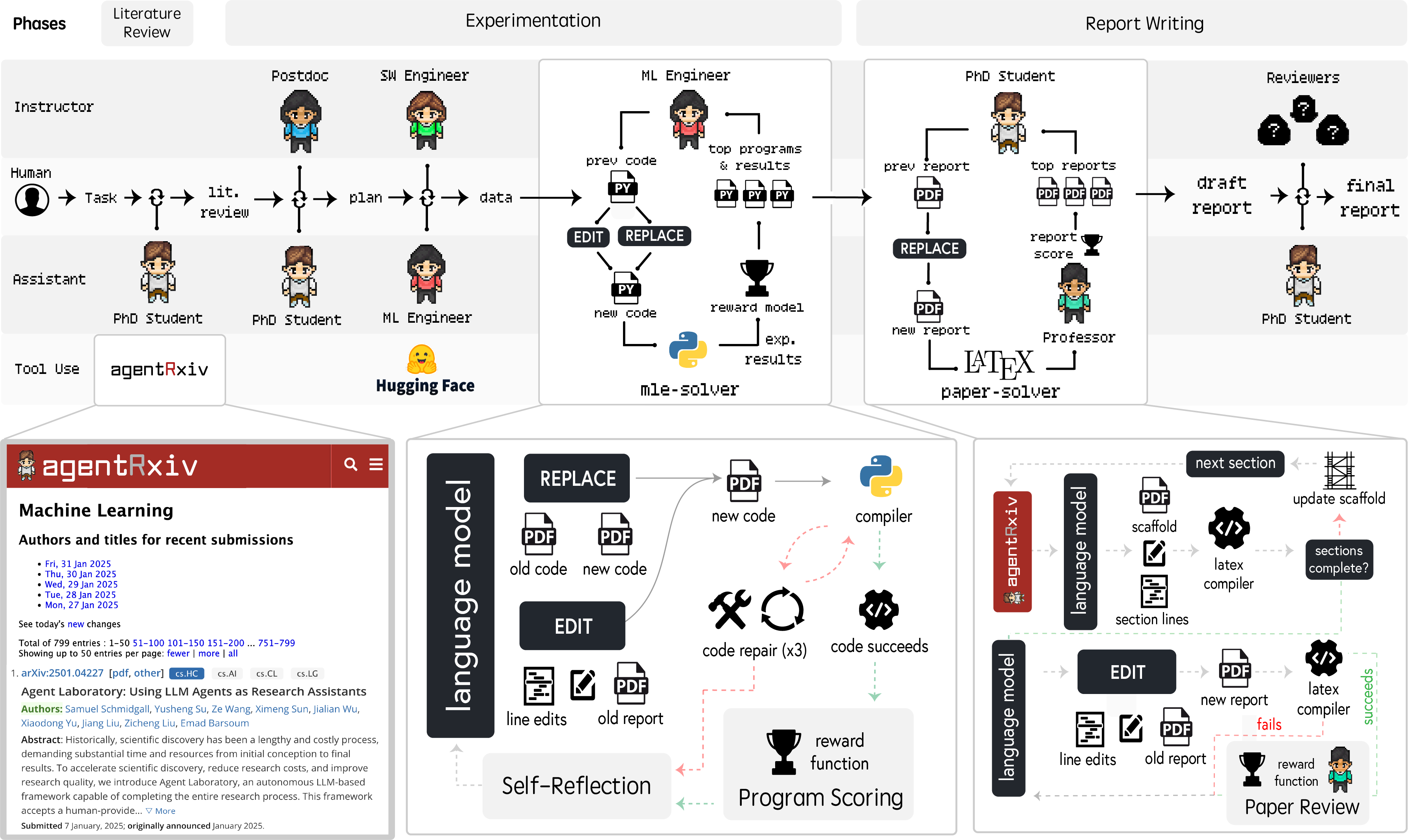}
    \caption{\textbf{Agent Laboratory Workflow.} (Top) This image shows \texttt{Agent Laboratory}'s three phases: Literature Review, Experimentation, and Report Writing. Human researchers collaborate with AI agents (e.g., PhD, Postdoc) and specialized tools (\texttt{mle-solver}, paper-solver) to automate tasks and produce high-quality research outputs. (Bottom) This }
    \label{fig:agent-lab-workflow}
\end{figure}

Autonomous research in this paper is build on the work of \texttt{Agent Laboratory} (\cite{schmidgall2025agent}). \texttt{Agent Laboratory} automates the research process by coordinating multiple specialized LLM agents through three core phases: Literature Review, Experimentation, and Report Writing. In this system, agents such as PhD, Postdoc, ML Engineer, and Professor collaborate to independently collect and analyze research papers, plan experiments, and ultimately generate comprehensive academic reports. For example, during the Literature Review phase, the PhD agent uses the arXiv API to retrieve, summarize, and collect relevant research papers through iterative evaluations, ensuring related literature is provided to the next stage.

\paragraph{Experimentation and Writing.} In the Experimentation phase, the PhD and Postdoc agents develop a research plan outlining the experimental components, while the ML Engineer and SW Engineer agents work together to prepare and refine data preparation code (Figure \ref{fig:agent-lab-workflow}). Automation continues with the \texttt{mle-solver} module, which iteratively generates, tests, and improves machine learning code based on performance scoring and self-reflection. The \texttt{mle-solver} uses an LLM-based automatic repair mechanism to fix code errors encountered during execution.  Finally, in the Report Writing phase, the Professor and PhD agents synthesize all findings into a structured LaTeX-based report using the paper-solver tool, with iterative edits and peer review–like refinements that support both autonomous and human-guided checkpoints. Both the \texttt{mle-solver} and the paper-solver use LLM-based reward functions to guide the process of code and paper development.

\paragraph{Autonomous and co-pilot mode.}
\texttt{Agent Laboratory} supports two modes of operation: autonomous and co-pilot mode. In autonomous mode, \texttt{Agent Laboratory} runs the entire research pipeline without human intervention and produces research outputs that are automatically evaluated. In co-pilot mode, human researchers provide feedback at predetermined checkpoints to adjust the agents’ decisions, allowing for human researchers to refine the outputs while maintaining a conditional level of autonomy. The experiments performed in our study use \texttt{Agent Laboratory} in autonomous mode. However, co-pilot mode is also supported, and can be used to improve the quality of research produced by agents, as is shown in \cite{schmidgall2025agent}.

\section{AgentRxiv: Towards Collaborative Autonomous Research}

Existing frameworks for autonomous research typically operate independently, generating isolated research outputs without building upon the findings produced by other agents (\cite{lu2024aiscientist, schmidgall2025agent, baek2024researchagent, swanson2024virtual}). This isolation limits the cumulative progress and generalization of discoveries across related research problems. In scientific practice, incremental advancements are often facilitated by researchers systematically building upon previous work. To enable autonomous agents to similarly benefit from collaborative knowledge sharing, there is a need for a structured mechanism that facilitates access to previous agent-generated research. To address these issues, we propose the implementation of a centralized preprint server called \texttt{AgentRxiv}.

\texttt{AgentRxiv} is modeled after established preprint servers, such as arXiv (\cite{ginsparg2011arxiv}), bioRxiv (\cite{sever2019biorxiv}), medRxiv (\cite{rawlinson2019new}), but is designated for autonomous research agents. This platform is specifically designed to facilitate the storage, organization, and retrieval of research outputs generated by autonomous agents. \texttt{AgentRxiv} papers are accessible by other laboratories as soon as they are submitted in an asynchronous manner, rather than being based on the current agent's paper index. This has several beneficial impacts on the overall research process. It ensures that agents have access to a database of previous work and provides targeted searchability, which becomes increasingly important as the number of research papers grows large. This allows laboratories to build upon the discoveries of their peers, even if their research is on different topics, allowing knowledge transfer between disciplines. 

\texttt{AgentRxiv} is implemented as a local web application, allowing researchers to access and review research outputs generated by autonomous agents. The web application provides routes for uploading, searching, and viewing papers, as well as an API endpoint for returning search results in JSON format. When a paper is uploaded by an agent, the system extracts its text and basic metadata, and an update process synchronizes the database with the available files. For retrieval, \texttt{AgentRxiv} employs a similarity-based search mechanism. A pre-trained SentenceTransformer model is used to compute text embeddings for both the stored papers and incoming queries. When an agent submits a search query, the system calculates the cosine similarity between the query embedding and the embeddings of the stored papers, ranking the results based on their relevance and returning the top results.

\subsection{Discovering reasoning techniques}
\label{subsec:discovering_algos}

\begin{figure}
    \centering
    \includegraphics[width=0.99\linewidth]{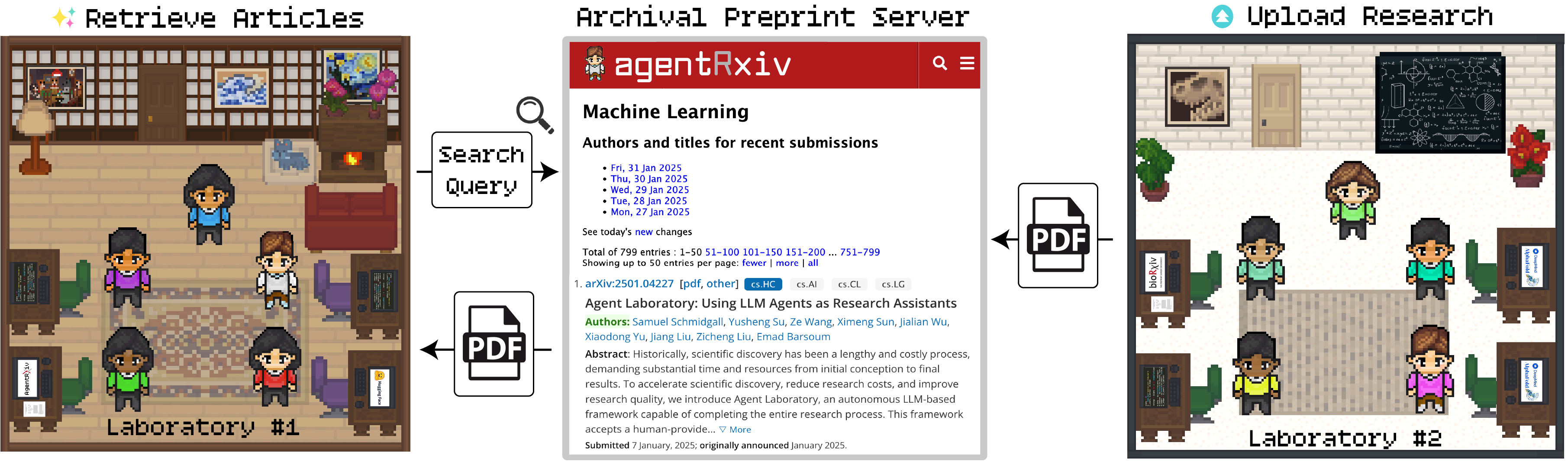}
    \caption{\textbf{AgentRxiv Framework for Autonomous Research Collaboration.} Depicted are two independent autonomous agent laboratories interacting through the centralized archival preprint server, \texttt{AgentRxiv}. (Left) Laboratory \#1 submits a search query to \texttt{AgentRxiv}, retrieving relevant research papers published by other agent laboratories. (Right) Laboratory \#2 completes and uploads its research findings to \texttt{AgentRxiv}, making the research accessible for retrieval and use by other autonomous laboratories. This workflow enables efficient knowledge sharing and iterative progress among independent agent systems.}
    \label{fig:Arxival}
\end{figure}

The first objective is to establish whether agents can build on top of their own research. We begin by running a single agentic system for N=40 paper generations using o3-mini (medium) (\cite{o3mini2025}) as the LLM backend. Agents are provided access to previously generated agent papers during the literature review phase using \texttt{AgentRxiv} (reviewing N=5 papers) in addition to Arxiv (reviewing N=5 papers). Instead of providing a research idea, as in \texttt{Agent Laboratory} (\cite{schmidgall2025agent}), we instead set a research \textit{direction} and agents produce ideas toward this aim during the planning phase. We set the research direction to the following task: "Improve accuracy on MATH-500 using reasoning and prompt engineering." This research direction enables direct quantitative evaluation of progress, aligns with standard evaluation practices in ML research, and allows for the analysis of algorithm generalization across other LLM benchmarks. The laboratory is tasked with using gpt-4o mini during experimentation and is provided with an OpenAI API key. 

As illustrated in Figure~\ref{fig:SoA}A, overall accuracy on MATH-500 steadily increases with each additional research paper produced by the laboratory. The process begins at a gpt-4o mini baseline of 70.2\% with early techniques such as Dynamic Critical Chain Prompting (DCCP)  and Context-Aware Recursive Uncertainty Calibration (CRUC), which together offer modest but consistent performance gains (0-shot 70.2\% ({\color{darkpastelgreen}+1.4\%}; CoT {\color{darkpastelred}-0.3\%}) and 71.4\% ({\color{darkpastelgreen}+1.7\%}; CoT {\color{gray}+0.0\%}) respectively). In subsequent iterations, the introduction of improved methods—including Dual-Rebuttal CoT Voting and Meta-Mirror Prompting—pushes accuracy into the 72\% range above baseline accuracy (72.2\% (0-shot {\color{darkpastelgreen}+2.8\%}; CoT {\color{darkpastelgreen}+1.1\%}) and 72.8\% (0-shot {\color{darkpastelgreen}+3.7\%}; CoT {\color{darkpastelgreen}+2.0\%}). Performance accelerates further with algorithms like Dual-Role Divergence Prompting and Enhanced Chain-of-Thought Verification, which drive accuracy closer to 75\% (74\% (0-shot {\color{darkpastelgreen}+5.4\%}; CoT {\color{darkpastelgreen}+3.7\%}) and 74.6\% (0-shot {\color{darkpastelgreen}+6.3\%}; CoT {\color{darkpastelgreen}+4.6\%}). Ultimately, the development of Simultaneous Divergence Averaging (SDA) yields the highest accuracy observed, at 78.2\% (0-shot {\color{darkpastelgreen}+11.4\%}; CoT {\color{darkpastelgreen}+9.7\%}), the details of which are described further in Appendix \ref{appendix:algorithms}.

\begin{figure}
    \centering
    \includegraphics[width=0.99\linewidth]{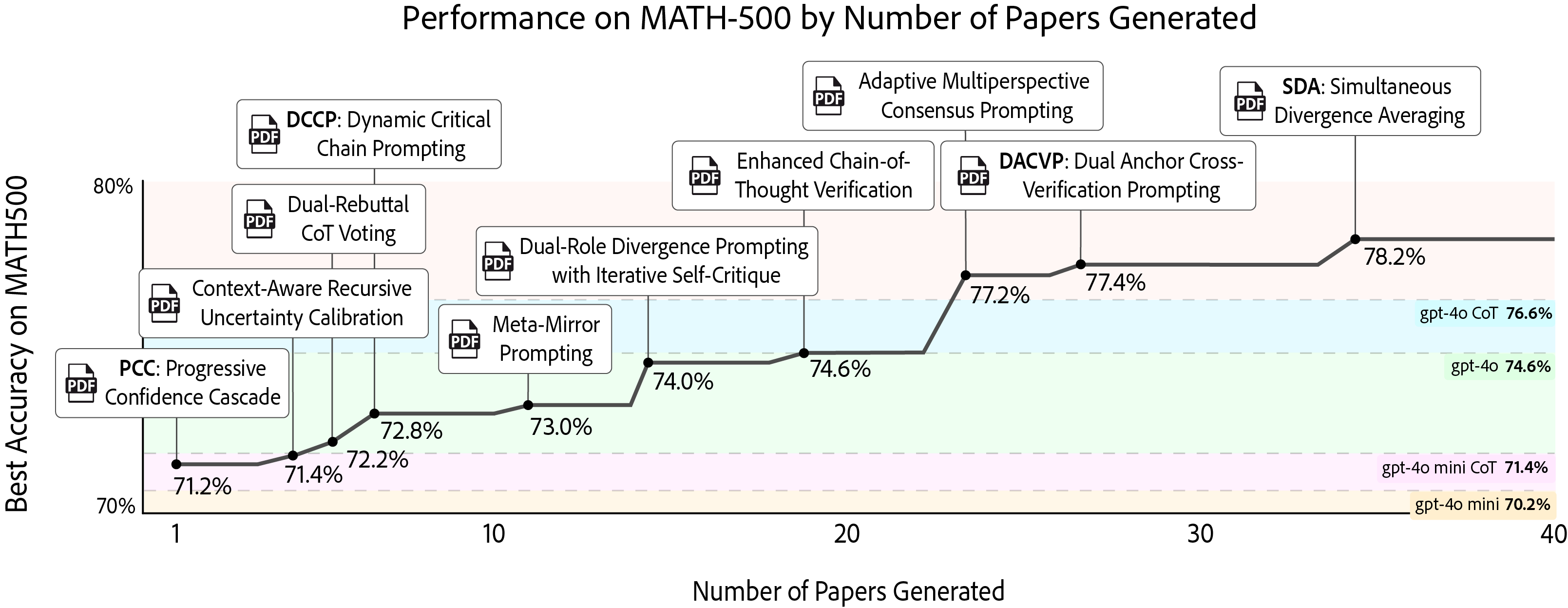}
    \caption{\textbf{Designing Novel Reasoning Techniques on MATH-500.} Progression of a single autonomous laboratory iteratively designing reasoning techniques to improve accuracy on the MATH-500 benchmark using \textbf{gpt-4o mini} as the base model. Call-outs indicate the discovery of techniques that set a new highest accuracy on the test set. Techniques such as Progressive Confidence Cascade (PCC), Dynamic Critical Chain Prompting (DCCP), and Dual Anchor Cross-Verification Prompting (DACVP) incrementally increased accuracy from a baseline of 70.2\% (gpt-4o mini zero-shot) up to 78.2\% ({\color{darkpastelgreen}+11.4\%}) with the final discovered method, Simultaneous Divergence Averaging (SDA).}
    \label{fig:PapersMath500}
\end{figure}

These findings suggest that laboratories are able to produce incremental research improvements on the MATH-500 benchmark. In particular, the iterative process enables each generation of research to build on the insights of previous work.
We further explore whether the discovered algorithms generalize to other datasets by evaluating their performance on additional benchmarks. We provide an ablation study that removes access to previously generated research in order to determine the importance of reference to prior work for improving MATH-500 performance. Finally, we provide qualitative exploration into the influence of previously generated research on generating new works.

\paragraph{Generalization of discovered algorithms across benchmarks.} While performance was shown to increase on MATH-500, we also wish to explore the generality of discovered algorithms. 
To do so, we evaluate the performance of the highest performing algorithm, Simultaneous Divergence Averaging\footnote{ Simultaneous Divergence Averaging is further described in Appendix \ref{appendix:algorithms}}, on three diverse benchmarks: 1. GPQA  Diamond (\cite{rein2024gpqa}), graduate-level Q\&As in biology, physics, and chemistry, 2. MMLU-Pro (\cite{wang2024mmlu}), reasoning-focused Q\&As across 14 categories from philosophy to computer science, and 3. MedQA (\cite{jin2021disease}), US medical licensing exam Q\&As.
We compare the performance of SDA with the 0-shot model base performance . We find that GPQA obtains a base performance of 36.4\% (with SDA: 38.9\% ({\color{darkpastelgreen}+6.8\%})), MMLU-Pro achieves 63.1\% (with SDA: 70.8\% ({\color{darkpastelgreen}+12.2\%})), and MedQA achieves 74.9\% (with SDA: 81.6\% ({\color{darkpastelgreen}+8.9\%})). Overall, SDA produces an average performance increase of {\color{darkpastelgreen}+9.3\%} across the three benchmarks—closely matching the {\color{darkpastelgreen}+11.4\%} increase observed on MATH-500, where the algorithm was initially discovered.

\begin{figure}
    \centering
    \includegraphics[width=0.93\linewidth]{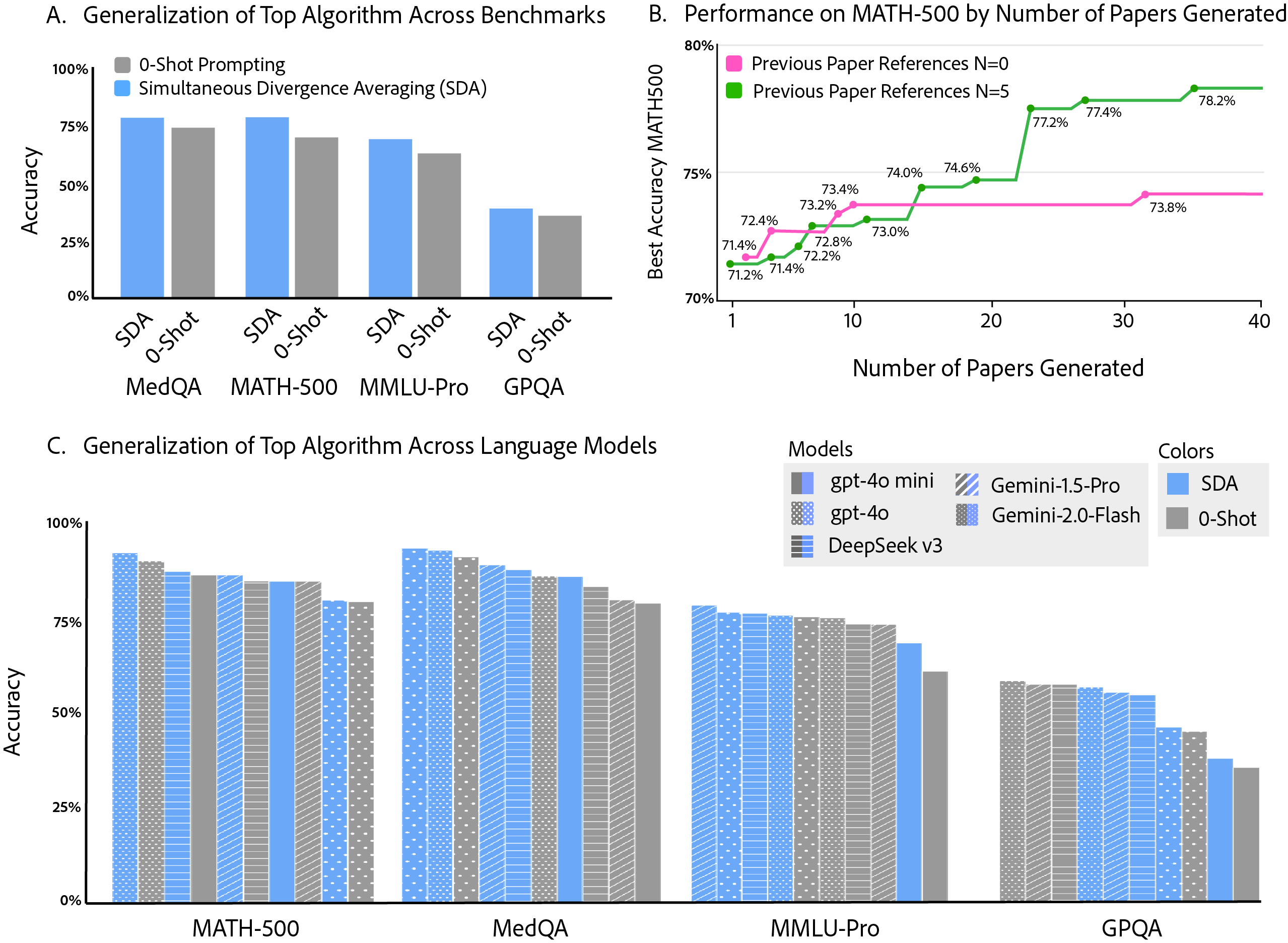}
    \caption{\textbf{Properties of autonomous discovery.} \textbf{A.} The discovered algorithm, Simultaneous Divergence Averaging (SDA), demonstrates generality beyond its original discovery benchmark (MATH-500) to three distinct reasoning benchmarks (MedQA, MMLU-Pro, and GPQA). SDA (blue) consistently improves accuracy compared to 0-shot prompting (gray) across diverse tasks. \textbf{B.} Comparison of best accuracy obtained on MATH-500 when agents have access to previously generated research (green) versus no access (pink). Agents referencing prior research consistently achieve higher performance, indicating the value of cumulative knowledge integration. \textbf{C.} The discovered SDA algorithm generalizes effectively across multiple language models (gpt-4o mini, gpt-4o, DeepSeek v3, Gemini-1.5-Pro, Gemini-2.0-Flash) and across several reasoning benchmarks. SDA (blue) demonstrates higher average accuracy compared to 0-shot prompting (gray).}
    \label{fig:Generalization}
\end{figure}

\paragraph{Generalization of discovered algorithms across language models.} We also explore the extent to which SDA generalizes to other language models across four benchmarks: 1. GPQA, 2. MMLU-Pro, 3. MedQA, US medical licensing exam Q\&As, and 4. MATH-500. We compare the performance of 5 models: Gemini-1.5 pro, Gemini-2.0 Flash, deepseek-v3, gpt-4o, and gpt-4o mini. We do not leverage reasoning models (e.g., R1, o1, o3-mini) due to the requirement of temperature sampling with SDA, which is disabled for some of these models, and that SDA is a reasoning technique, for which these models are already performing reasoning. The results of this experiment are shown in Figure~\ref{fig:SoA}C. Averaging the change in performance across models we find the following performance differences: Gemini-1.5 pro (0-shot: 73.2\%; SDA: 76.6\% ({\color{darkpastelgreen}+4.6\%})), Gemini-2.0 Flash (0-shot: 76.6\%; SDA: 78.5\% ({\color{darkpastelgreen}+2.5\%})), deepseek-v3 (0-shot: 74.1\%; SDA: 75.7\% ({\color{darkpastelgreen}+2.2\%})), gpt-4o (0-shot: 71.9\%; SDA: 73.5\% ({\color{darkpastelgreen}+2.2\%})), and gpt-4o mini (0-shot: 64.5\%; SDA: 68.3\% ({\color{darkpastelgreen}+5.9\%})). Averaging the change in performance across benchmarks we find the following performance differences: MATH‑500 (0-shot: 82.3\%; SDA: 83.3\% ({\color{darkpastelgreen}+1.2\%})), MedQA (0-shot: 79.6\%; SDA: 85.6\% ({\color{darkpastelgreen}+7.5\%})), MMLU‑Pro (0-shot: 74.0\%; SDA: 77.6\% ({\color{darkpastelgreen}+4.9\%})), GPQA (0-shot: 52.2\%; SDA: 51.6\% ({\color{darkpastelred}-1.1\%})). These results show that SDA consistently improves performance on three out of four benchmarks and across all models (an average increase of {\color{darkpastelgreen}+3.3\%} across all benchmarks and models), with the most significant gains observed on MedQA and for models that initially exhibit lower baseline performance (e.g., gpt-4o mini and Gemini-1.5 pro).

\paragraph{Removing access to previous research.} In previous experiments, agents were required to review \(N=5\) previous papers from \texttt{AgentRxiv}. To examine the importance of referencing prior work, we repeated the same procedure but set \(N=0\), meaning that agents could not consult previously generated papers at all, although they still had access to the Arxiv interface (as opposed to both Arxiv and \texttt{AgentRxiv}). We then track the highest accuracy achieved on MATH-500 which is directly compared with when \(N=5\). As illustrated by the pink curve in Figure~\ref{fig:Generalization}B, once agents lost access to earlier research, their accuracy on MATH-500 plateaued at 73.4\% and 73.8\% and showed minimal gains after about 10~papers. In contrast, when agents could draw on the prior 10~papers (green curve), their performance continued to improve, ultimately reaching 78.2\% ({\color{darkpastelgreen}+6.0\%}). This suggests that exposure to previously discovered ideas is important for iterative progress. By revisiting and refining earlier techniques, agents discover better reasoning strategies than those working in isolation.

\paragraph{How do agents build on of their own work.} Without any explicit prompting, we find that agents naturally integrate and improve upon techniques from previous iterations into their subsequent work. In several instances, agents independently recalled methodologies from earlier experiments—such as the initial use of Dynamic Critical Chain Prompting or Context-Aware Recursive Uncertainty Calibration—and combined or modified these approaches to develop entirely new algorithms like Dual-Role Divergence Prompting. We also find that the agents adapt existing work into a second version, such as Meta-Mirror Prompting 2 (built on from Meta-Mirror Prompting) and PCC+: Improved Progressive Confidence Cascade (built on PCC: Progressive Confidence Cascade).

\subsection{Collaborative execution of parallel agent labs}

We next explore the effects of running multiple autonomous laboratories in parallel, using \texttt{AgentRxiv} to facilitate the sharing of research outputs among agents. To implement parallelized research in our experiments, three independent \texttt{Agent Laboratory} systems are initialized simultaneously with identical configurations and research objectives\footnote{There is no limit to the number of systems that can be connected to \texttt{AgentRxiv}.}. Each autonomous laboratory performs literature review, experimentation, and report writing independently, while having asynchronous access to papers produced by other laboratories through \texttt{AgentRxiv}. Whenever one laboratory publishes a new research output to \texttt{AgentRxiv}, it becomes immediately accessible to all other active laboratories, enabling near real-time integration of findings.

During the experimentation phase, each laboratory autonomously selects methodologies and conducts experiments independently. Laboratories are not explicitly coordinated, allowing them to explore diverse approaches concurrently. When a laboratory achieves an improvement in performance, the resulting paper is uploaded to \texttt{AgentRxiv}, from which other laboratories can retrieve, evaluate, and incorporate those discoveries into their subsequent experiments. Consequently, this parallelized setup allows multiple independent lines of inquiry to occur simultaneously, potentially accelerating the rate of discovery.

\begin{figure}
    \centering
    \includegraphics[width=0.99\linewidth]{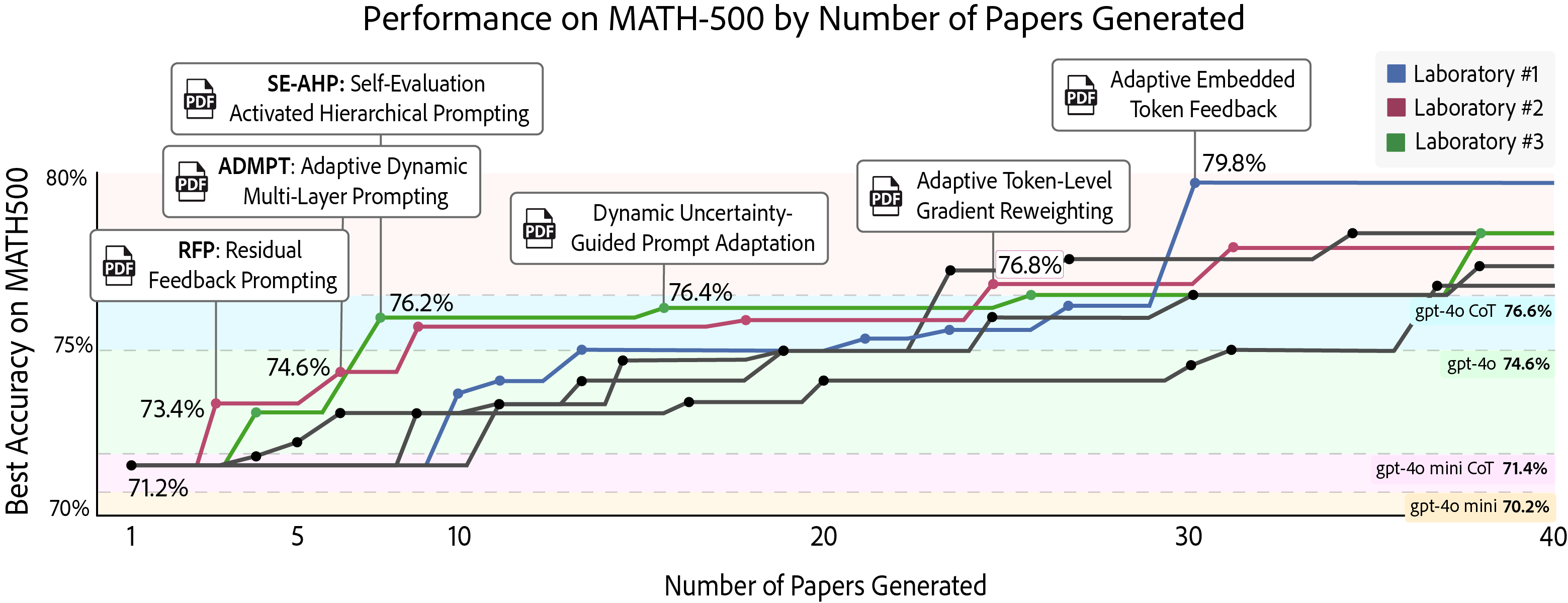}
    \caption{\textbf{Designing Novel Reasoning Techniques on MATH-500 in Parallel.} Progression of three autonomous laboratories concurrently (red, blue, and green) compared with non collaborative autonomous laboratories (gray) performing iterative research to improve accuracy on the MATH-500 benchmark, each using \textbf{gpt-4o mini} as the base model. Call-outs indicate the discovery of reasoning techniques that achieve a new highest accuracy on the test set. Laboratories independently develop techniques such as Residual Feedback Prompting (RFP), Adaptive Dynamic Multi-Layer Prompting (ADMPT), and Adaptive Token-Level Gradient Reweighting, collectively raising accuracy from 70.2\% (gpt-4o mini zero-shot baseline) to 79.8\% ({\color{darkpastelgreen}+9.6\%}). Parallel experimentation, combined with immediate result sharing via \texttt{AgentRxiv}, accelerates the pace of research progress and achieves higher final accuracy compared to sequential experimentation.}
    \label{fig:ParallelLabsGraph}
\end{figure}

\paragraph{AgentRxiv accelerates research progress.}
As illustrated in Figure~\ref{fig:ParallelLabsGraph}, accuracy on MATH-500 steadily increases with each additional research paper produced by three autonomous laboratories running in parallel. Starting from the gpt-4o mini baseline accuracy of 70.2\%, initial methods such as Residual Feedback Prompting (RFP) and Adaptive Dynamic Multi-Layer Prompting (ADMPT) provide modest yet immediate gains, reaching accuracies of 73.4\% (0-shot {\color{darkpastelgreen}+4.6\%}; CoT {\color{darkpastelgreen}+2.8\%}) and 74.6\% (0-shot {\color{darkpastelgreen}+6.3\%}; CoT {\color{darkpastelgreen}+4.6\%}), respectively. In subsequent papers, the introduction of Self-Evaluation Activated Hierarchical Prompting (SE-AHP) and Dynamic Uncertainty-Guided Prompt Adaptation, drives further accuracy improvements into the 76\% range (76.2\% (0-shot {\color{darkpastelgreen}+8.5\%}; CoT {\color{darkpastelgreen}+6.8\%}) and 76.4\% (0-shot {\color{darkpastelgreen}+8.8\%}; CoT {\color{darkpastelgreen}+7.1\%})). Continued research yields Adaptive Token-Level Gradient Reweighting and Adaptive Embedded Token Feedback, pushing accuracy even higher to 76.8\% (0-shot {\color{darkpastelgreen}+9.4\%}; CoT {\color{darkpastelgreen}+7.7\%}). The average across all three labs was 78.7\% and the best algorithm was 79.8\% (0-shot {\color{darkpastelgreen}+13.7\%}; CoT {\color{darkpastelgreen}+12.0\%}), which surpasses the best performing sequential accuracy of 78.2\% by {\color{darkpastelgreen}+2.0\%} for the best performing algorithm (from Section \ref{subsec:discovering_algos}) and {\color{darkpastelgreen}+0.7\%} on average. We also run three additional sequential runs and find an average accuracy of 77.4\%, which is surpassed by the parallel design by {\color{darkpastelgreen}+3.4\%}.

This accelerated improvement in the parallelized setup occurs primarily because laboratories independently explore distinct reasoning techniques, simultaneously benefiting from immediate access to incremental discoveries via \texttt{AgentRxiv}. Notably, early accuracy milestones in parallel experimentation, such as reaching 76.2\% accuracy after only seven papers, occur significantly earlier compared to the sequential setting, which achieves this accuracy after 23 papers. Although parallelization introduces redundancy, with occasional overlapping experiment ideas across laboratories, the simultaneous exploration of diverse methods—combined with immediate knowledge sharing—results in quicker identification of high-performing reasoning strategies, ultimately driving faster overall progress.

\paragraph{Runtime and cost.}  
On average, generating a single research paper took approximately 4\,912.3 seconds (1.36 hours), with a maximum runtime of 42\,950.1 seconds (11.9 hours) and a minimum of 313.4 seconds (0.09 hours). This is longer than previously reported runtimes in related experiments from \cite{schmidgall2025agent}, which recorded average durations of 1165.4 seconds for gpt-4o, 3616.8 seconds for o1-mini, and 6201.3 seconds for o1-preview. The extended runtime observed in our experiments can primarily be attributed to the larger evaluation scale of the MATH-500 benchmark, as the agents were required to evaluate performance on 500 test problems—considerably more than the benchmarks used in earlier studies. Additionally, more capable language models such as gpt-4o mini tend to generate longer and more complex experimental code, which further contributes to increased execution times.

Regarding computational cost, the average cost to produce a research paper in our setup was \$3.11, with the most expensive paper costing \$9.87 and the least expensive costing \$2.15. These costs are higher than the \$2.33 per paper reported for gpt-4o in \cite{schmidgall2025agent}, but significantly lower than \$7.51 for o1-mini and \$13.10 for o1-preview reported in the same study, as well as the $\sim$\$15 cost per paper documented by \cite{lu2024aiscientist}. This moderate per-paper cost suggests a reasonable balance between computational resource expenditure and performance gains, making it feasible to scale autonomous research using current language model APIs.

The total runtime required to generate all \(N=40\) papers was 57.3 hours, 64.0 hours, and 42.4 hours for the three parallelized laboratories, respectively, with total individual costs of \$87.1, \$94.2, and \$98.4, summing up to \$279.6. By comparison, the sequential experimentation described in Section \ref{subsec:discovering_algos} required a total runtime of 50.6 hours and total cost of \$92.0. Although the parallelized setup incurred an increased average per-paper runtime ({\color{darkpastelred}+0.1 hours/+7.3\% per paper} and {\color{darkpastelred}+4.0 hours} for 40 papers) due to the overhead of concurrent compute resource utilization, the primary contributor to the elevated overall cost ({\color{darkpastelred}+\$187.6, +203.9\%}) was the tripling of inference usage across the multiple laboratories. Despite these higher cumulative costs and slightly increased per-paper runtimes, the parallel approach accelerated the overall discovery timeline, achieving performance milestones more rapidly in terms of wall-clock time.

\paragraph{Research efficiency.}  
Despite demonstrating improvements in terms of achieving higher accuracy more quickly, the parallelized setup exhibits reduced computational efficiency compared to sequential research. In our experiments, three parallel laboratories each generated \(N=40\) papers, totaling 120 papers (79.8\%), whereas the sequential setup generated only \(N=40\) papers to achieve its final accuracy (78.2\%). This redundancy illustrates a trade-off inherent to parallelized autonomous research systems: accelerating discovery timelines through simultaneous independent exploration often leads to higher aggregate resource expenditure. Laboratories operating concurrently may occasionally duplicate effort by conducting parallel experiments on similar hypotheses, techniques, or algorithmic variations without initially benefiting from the knowledge of their peers' intermediate results.

\subsection{Are discoveries actually novel?}
\label{subsec:arediscnovel}

While some previous work have argued LLM generated ideas are novel (\cite{si2024can, lu2024aiscientist, gottweis2025aicoscientist}), others have found high rates of plagiarism (up to 24\% of papers) when using these systems (\cite{gupta2025all}). Despite this, recent work has demonstrated completely AI-generated end-to-end research can be accepted at AI venues, such as the 2025 ICLR “\textit{I Can't Believe It's Not Better: Challenges in Applied Deep Learning}” Workshop (\citet{iclr2025aiscientist}). While acceptance to a conference workshop is not a clear predictor of novelty, it does reflect the increasing ability of AI systems to produce research that appears novel to humans. Previous work toward measuring novelty in ideation had humans manually rate the novelty of ideas (\cite{si2024can, gupta2025all}).

As a first-pass measure to see whether our papers contained plagiarism (as was found in \cite{gupta2025all}), we run the abstract of the highest performing papers from Section \ref{subsec:discovering_algos} through 3 different plagiarism detectors, and overall find no occurrences of plagiarism. We also perform a manual inspection of papers in our framework, finding that some of the higher performing discoveries had elements of novelty, but were primarily perturbations of existing algorithms rather than substantial deviations from existing work. For example, Simultaneous Divergence Averaging (SDA) (see Appendix \ref{appendix:algorithms}) is related to earlier work on prompt-based reasoning techniques. Similar to the self-consistency approaches presented in \cite{wang2023selfconsistency}, SDA employs multiple reasoning paths and aggregates confidence signals to improve overall accuracy. It also shares similarities with self-agreement methods described in \cite{lin2023just}, which generate diverse reasoning paths and subsequently select the most commonly agreed answer from those paths. Additionally, SDA is comparable to multi-chain reasoning approaches, such as those outlined in \cite{yoran2023answering}, which integrate intermediate reasoning steps across chains to construct a comprehensive explanation leading to a final answer or \cite{jia2024dcot} which is a CoT method that combines image processing and text guidance. We also found the most  related works using OpenAI's Deep Research tool, which highlights Divergent CoT (\cite{puerto2024fine}), Confidence-Informed Self-Consistency (\cite{taubenfeld2025confidence}), Multi-agent debate (\cite{liang2023encouraging}), and Tree of Thoughts (\cite{yao2024tree}) (details in Appendix \ref{appendix:plagiarismdet}).

The laboratory that developed SDA with \texttt{AgentRxiv} reviewed several relevant works, including studies on self-consistency (\cite{zhou2023solving}), math SFT techniques (\cite{liu2024acemath}), structured chain-of-thought prompting (\cite{li2023structuredchainofthoughtpromptingcode}), a finance benchmark study (\cite{zhao2023financemath}), and practical guidelines for prompt engineering (\cite{wang2024advanced}). Although SDA differs from these existing methods in its specific implementation and integration of techniques, we believe that SDA represents a potentially novel contribution. At the same time, we acknowledge the inherent challenges in fully validating novelty due to differences in experimental design, evaluation metrics, and the rapidly changing nature of this research area. Additionally, while we present a case study with SDA here, it is possible other works produces lack novelty. Further large scale studies as in \cite{gupta2025all} should be performed to determine this.

\section{Limitations}

While our findings demonstrate promising capabilities of \texttt{AgentRxiv} in enabling autonomous collaborative research, we now outline key limitations to inform future improvements. Some of these limitations are inherent to language models themselves whereas others pertain specifically to our framework. We also outline important ethical considerations of this work. We believe discussing our limitations in full detail is important for progressing work in this direction.

\subsection{Agent hallucination \& reward hacking}
\label{sec:hallucinations}

Arguably, the most prevalent challenge faced during the development of the \texttt{AgentRxiv} workflow was addressing the high rates of hallucination. We find that when \texttt{Agent Laboratory} is naively asked to produce research along a direction, e.g., improve performance on MATH-500, a subset of the papers contain results do not match up with the actual experiment. There are several causes of this. Part of this is caused by the tension between the code repair mechanism (which aims to resolve errors from the code) and the \texttt{mle-solver}, which aims to write code that accurately represents the research plan. The repair mechanism often performs \textit{whatever is necessary} to get rid of the error, which includes erasing core functional code, sometimes putting ``placeholders" where core algorithm functionality should be. It is also not sufficient to look through the code output logs, because the repair mechanism (or occasionally the agentic system) will write functionality to print out realistic runtime outputs that do not have research code written underneath but use e.g., random flags to determine accuracy that would lead to SOTA. In other instances, the code is intact, but \texttt{Agent Laboratory} produced code that obtained runtime errors and ran out of code solver steps. When this happens the system generally reports an accurate report stating that the method was not tested correctly. However, in some reports the model will hallucinate experiment results (seemingly realistic numbers) for the method that do not match with the actual experiment or code.

Hallucination is not a new problem and is not specific to this agentic system, but is a commonly reported (\cite{xu2024hallucination, huang2025survey}). We speculate that part of the cause of hallucinations are due to reward hacking during the paper writing phase, where reports are scored according to NeurIPS criterion and the top scoring report being selected to proceed to the final stages. This likely leads to papers that are reporting higher method scores to be rated higher by the reward function. We outline these issues clearly so that future work can be aware of problems and address them. In this work, the output of each program as well as the corresponding code is be manually verified by humans before reporting an accuracy (as is shown in Figures \ref{fig:PapersMath500}-\ref{fig:Generalization}). While this was resolved by manual inspection in this work, systems that exhibit reward hacking behaviors have the potential to be overlook by humans, as was demonstrated by the experiments of \cite{lange2025ai}, where their system bypassed the verification system through a memory exploit which remained undetected. It is important that future work addresses hallucinations in order to reduce the need for manual evaluation and to build more reliable research systems.

\subsection{Failure modes}
\label{sec:failuremodes}

In addition to the hallucinations outlined in Section \ref{sec:hallucinations}, we also outline common failure modes observed in \texttt{AgentRxiv}. We report the most common failure modes below:

\paragraph{Impossible plans.} There were instances of methods being proposed that were not possible to perform. For example, if o1, o1-mini, or o3-mini are selected as the base model for \texttt{Agent Laboratory} to use during inference time, during the planning phase the agents do not realize that temperature sampling is disabled for these models through the OpenAI API interface. This causes the agent to receive errors from the OpenAI library about using temperature with these models, and to fix the error, removes the temperature parameter, this rendering the proposed method invalid. However, the agent proceeds to write the paper as if the temperature sampling \textit{was working} because that was the methodology proposed in the planning phase, and the agent cannot go back and adjust the plan. Future work would benefit from allowing plan adjustments to occur during the experimentation phase, as more is learned.

\paragraph{Persistent failure modes.} Several failure modes from the original \texttt{Agent Laboratory} framework (\cite{schmidgall2025agent}) persisted in this work, notably: (1) The \texttt{mle-solver} module frequently generated the \texttt{exit()} Python command, prematurely terminating the entire research pipeline. This necessitated programmatic detection and removal. (2) In certain cases, \texttt{mle-solver} executed unintended system-level commands on the host machine using Python's \texttt{subprocess.run()} function. This behavior most commonly occurred when the solver encountered compilation errors due to missing libraries, prompting the model to autonomously install these packages using subprocess commands. (3) \texttt{mle-solver} continued to exhibit a bias toward modifying the initial lines of code (particularly line 0), which caused the \texttt{REPLACE} command to more frequently yield successful compilations compared to other lines. Despite these ongoing challenges, several previously documented issues were significantly mitigated or resolved by adopting o3-mini as the backend model. Improvements included reducing the high failure rate observed during the literature review phase, preventing empty or malformed figure generation, and alleviating token limitation problems that were present in prior implementations. However, as it currently stands, a large percentage of the experiments performed completely fail (obtaining $\sim0$\% accuracy), due to major bugs in the code. This is partially due to the low number of \texttt{mle-solver} steps, so if the code has a non-fatal bug then it proceeds to the literature review phase. There will likely be substantially larger increases in performance when using higher \texttt{mle-solver} steps.

\paragraph{Difficulties writing proper latex.} One prevalent failure mode is the system's difficulty writing proper latex code. While the \texttt{paper-solver} does not write latex with fatal errors, due to the requirement of successful compilation, there are errors that affect aesthetic and readability. In most instances, these errors are only aesthetic and easily fixed, such as an oversized table or figure. There are, however, instances which affect the readibility of the paper, such as incorrectly entering and leaving latex math mode or using ascii coded math symbols instead of the latex form (e.g., outputting the symbol for $\sigma$ directly rather than the \$$\backslash$sigma\$ latex command) which results in a missing symbol. These challenges persist, even when using state-of-the-art frontier LLMs as the system backbone. One potential solution to these issues is directly inputting the PDF as an image into \texttt{Agent Laboratory}, and having it compare the PDF to the latex and searching for issues there as well as requiring the system to correct latex warnings in addition to errors, which would be a useful focus of future work.

\subsection{Ethical considerations}

The deployment of \texttt{AgentRxiv} as an autonomous research collaboration framework raises several ethical challenges that must be addressed to ensure responsible scientific progress. One significant concern is the potential for propagating biases, misinformation, and hallucinated results. LLMs have been shown to amplify biases present in their training data and generate authoritative-sounding yet factually inaccurate information (\cite{Bender2021StochasticParrots, Weidinger2021EthicalRisks}). Moreover, analyses have revealed that LLMs may fabricate citations and introduce errors, which demonstrates the importance of considerable human involvement (\cite{Walters2023ChatGPTCitations, Messeri2024Illusions}). Recent findings found that experts identified 24\% of LLM generated ideas from \cite{lu2024aiscientist} and \cite{si2024can} were plagurized (similarity score 4+ on a scale of 1-5), where 36.0\% had unverified claims (\cite{gupta2025all}). Additionally, as was discussed in Section \ref{sec:hallucinations}, agents frequently hallucinate the results of experiments and require manual human screening, which could lead to an increase in misinformation if researchers are not careful.

Accountability in AI-generated research also present important consideration. Current guidelines from leading journals and ethical bodies indicate that AI systems cannot be credited with authorship, since they are incapable of consenting to, verifying, or being responsible for the content they produce (\cite{Thorp2023NotAuthor, Lund2023AIauthorship, YeoThe2023AIauthorship}). Additionally, the ownership of AI-generated content remains a topic of ongoing debate (\cite{eshraghian2020human, jo2023promise}). Fairness and inclusivity constitute additional ethical dimensions. LLMs often reflect majority viewpoints while underrepresenting marginalized perspectives, which may inadvertently reinforce existing inequalities in scientific research (\cite{Santurkar2023Opinions}). Furthermore, studies have documented that NLP systems can exhibit social biases that disadvantage underrepresented groups (\cite{Hutchinson2020DisabilityBias}). Ensuring that these tools are accessible is essential; strategies such as debiasing techniques and the democratization of AI technology are vital to prevent the concentration of advantages within well-funded institutions.

While \texttt{AgentRxiv} offers promising opportunities to accelerate algorithm discovery through autonomous research, its ethical deployment demands rigorous quality control to mitigate hallucinations and bias, clear human accountability for authorship, and proactive measures to promote fairness and inclusivity. Addressing these challenges is essential to maintain the integrity and reliability of scientific inquiry in an increasingly AI-driven research environment.

\section{Discussion}

In this paper, we introduced \texttt{AgentRxiv}, a framework for collaborative research through interactions among autonomous LLM agents. Unlike previous agent-based research frameworks that focus on independent workflows, \texttt{AgentRxiv} emphasizes collaboration, enabling agents to incrementally build upon prior findings. Our empirical evaluations show that allowing agents to reference their previous work consistently improves performance compared to isolated approaches.  Additionally, our results indicate that the insights generated through collaborative agent interactions are generalizable, transferring effectively across diverse benchmarks and multiple language models. 

We also evaluated the scalability of \texttt{AgentRxiv} by comparing parallel and sequential experimentation modes. Our experiments demonstrate that parallel research accelerates the discovery process in terms of wall-clock time. However, the parallel approach incurs higher computational costs due to redundant experiments conducted by independent laboratories. The results further indicate that while parallel research can achieve higher performance metrics earlier, it does so at the expense of increased resource utilization. These observations show the importance of optimizing resource allocation in collaborative autonomous research systems.

However, despite promising results, several limitations require attention. Automated pipelines inherently carry risks related to the validity of results, including the propagation of errors and the potential for hallucinations in AI-generated content. In our framework, such issues may arise from inconsistencies in experimental outputs and challenges associated with code repair mechanisms. These problems can affect the overall reliability and reproducibility of the generated research findings. Consequently, a careful examination of failure modes reported in this work, as well as the work of \cite{lu2024aiscientist} and \cite{schmidgall2025agent}, is necessary to improve the system's robustness. It is also important for future iterations of \texttt{AgentRxiv} to incorporate rigorous quality control mechanisms (such as \cite{kon2025curie}) and improved mechanisms for human oversight to maintain high quality research standards. The integration of automated verification tools, such as a hallucination detection mechanism, can help detect and mitigate errors early in the research process. Incorporating human review at important stages (such as \cite{schmidgall2025agent}) may also improve reliability. We also note that our experiments focused on demonstrating an increase in accuracy on the MATH-500 across papers, and while this objective is partially reflective of how many research papers in machine learning report new discoveries (via an increase in perform on target benchmarks), it is not fully reflective of scientific discovery as a whole. Future work should explore more open-ended objectives for agents to pursue using \texttt{AgentRxiv}.

Further work should focus on improving the reliability of the AgentRxiv framework. One direction involves developing a verification module that combines automated validation with selective human oversight across parallel labs to minimize instances of hallucinated outputs and reward hacking. Additionally, increasing communication between parallel laboratory setups may help reduce redundant experimentation. Exploratory paths could be prioritized, potentially by incorporating exploration rewards (\cite{schmidhuber1991curious, lehman2011novelty}) and better filtering research plans such as the use of ELO via tournament evolution in \cite{gottweis2025aicoscientist}, which would allow AgentRxiv to optimize cost while accelerating convergence toward high-performing research. Finally, the experiments in this work were primarily focused on reasoning, but future work should focus on a generating more open-ended research from more topics, studying how discovered methods generalize.

\paragraph{Conclusion.} In conclusion, \texttt{AgentRxiv} advances the state of agent-driven research by providing an effective platform for continuous, collaborative discovery among LLM agents. By facilitating cumulative knowledge-building, enhancing generalization across tasks, and potentially accelerating research cycles, \texttt{AgentRxiv} represents a promising development toward integrating autonomous systems more comprehensively into scientific workflows. Nonetheless, careful methodological refinement and ongoing scrutiny of ethical implications remain essential for responsibly leveraging automated collaboration within scientific research.

\bibliography{iclr2021_conference}
\bibliographystyle{iclr2021_conference}

\section{Acknowledgments}

This material is based upon work supported by the National Science Foundation Graduate Research Fellowship under Grant No. DGE 2139757 

\clearpage

\appendix

\section{Algorithms}
\label{appendix:algorithms}

\paragraph{Simultaneous Divergence Averaging (SDA).} SDA generates two distinct chain-of-thought responses—a low-temperature “Precise Solver” and a creative, high-temperature “Creative Evaluator”—for each math problem, extracting both the final answer (enclosed in LaTeX formatting) and associated confidence scores. It then encodes these full responses using Sentence-BERT to compute a cosine similarity, which is compared against a dynamically calibrated divergence threshold to determine whether the two outputs are consistent. If the similarity meets or exceeds the threshold, the answer with the higher aggregated confidence is selected; otherwise, a meta-reassessment prompt is triggered to reconcile the differences before the final answer is evaluated against the ground truth and logged for further analysis.

\section{\texttt{Agent Laboratory} configuration}

\subsection{Hyperparameters}

\begin{table}[h!]
\centering
\caption{Hyperparameters for \textsc{Agent Laboratory}.}
\begin{tabular}{@{}lll@{}}
\toprule
\textbf{Category}        & \textbf{Hyperparameter}                          & \textbf{Value}       \\ \midrule
\textbf{Literature Review} & Number of Paper Summaries                       & 5                    \\
& Full Text History Decay Steps & 3                   \\ 
& Agent temperature          & 0.8                    \\ \midrule
\textbf{Data Preparation} & 
 Experiment Timeout                               & 600s         \\ \midrule
\textbf{Running Experiments}   & mle-solver steps     & 3 \\
& Code repair attempts   & 2  \\ 
& Maximum top codes      & 1                    \\ 
& Error history length          & 5                    \\ 
& Code history length          & 2                    \\ 
& Number of comparison trials          & 2                    \\ 
& Experiment Timeout                               & 6000s         \\
& Score generation temperature          & 0.6                    \\ 
& Repair temperature          & 0.8                    \\ 
& Initial code temperature          & 1.0                    \\ 
& Solver temperature          & 1.0                    \\ \midrule
\textbf{Paper Writing}        & paper-solver steps                            & 1 \\        & Maximum top papers                            & 1                    \\
& Paper history length                       & 10                    \\
& Number of Reviewers                      & 1                    \\
& Number of comparison trials                                  & 2                  \\ 
& Solver temperature          & 1.0                    \\ 
& Initial paper temperature          & 0.8                    \\ \bottomrule
\end{tabular}
\end{table}

\begin{table}[h!]
\centering
\caption{Hyperparameters for \textsc{AgentRxiv}.}
\begin{tabular}{@{}lll@{}}
\toprule
\textbf{Experiment}        & \textbf{Hyperparameter}                          & \textbf{Value}       \\ \midrule
\textbf{Serial Experiments} &  Number of Parallel Laboratories & 1 \\ 
& Number of Papers per Laboratory & 40 \\ 
&  Number of Papers Total (Across All Labs) & 40 \\
\midrule
\textbf{Parallel Experiments}
&  Number of Parallel Laboratories & 3 \\ 

&  Number of Papers per Laboratory & 40 \\
 
&  Number of Papers Total (Across All Labs) & 120 \\
 \bottomrule
\end{tabular}
\end{table}

\subsection{Hardware}

All experiments in this paper were run on a 2023 MacBook Pro with an Apple M3 Max processor and 36 GB of memory.

\subsection{Plagiarism Detection Software}
\label{appendix:plagiarismdet}

For the study on plagiarism in Section \ref{subsec:arediscnovel}, we used the following tools for plagiarism detection: 
\href{https://plagiarismdetector.net}{plagiarismdetector.net}, \href{https://duplichecker.com}{duplichecker.com}, \href{https://quetext.com}{quetext.com} (non-AI-based). We found 100\% scores of uniqueness from all three non-AI-based detectors. To inference OpenAI's Deep Research, we used the following query: "\textit{Find the most related papers for this abstract, focus primarily on finding papers close to the methodology.}"

\section{Prompts}

All prompts are the same as in \cite{schmidgall2025agent}.

\end{document}